\documentclass{article}
\usepackage{spconf,amsmath,graphicx}
\usepackage{url}
\usepackage{graphicx}
\usepackage{multirow}
\usepackage{cleveref}

\usepackage{graphicx}
\usepackage{tabularx}
\usepackage{color}

\title{Stacked Neural Networks for end-to-end ciliary motion analysis}
\begin{document}
\name{Charles Lu$^{\star}$ \qquad M. Marx$^{\dagger}$  \qquad M. Zahid$^{\S}$ \qquad C. W. Lo$^{\S}$ \qquad C. Chennubhotla$^{\dagger}$ \qquad S. P. Quinn$^{\star \ddag}$ }

 \address{$^{\star}$ Department of Computer Science, University of Georgia, Athens, GA USA \\
     $^{\dagger}$ Department of Computational and Systems Biology, University of Pittsburgh, Pittsburgh, PA USA \\
     $^{\S}$ Department of Developmental Biology, University of Pittsburgh Medical Center, Pittsburgh, PA USA \\
     $^{\ddag}$ Corresponding author \{\textrm{squinn@cs.uga.edu}\}}
%

%
\maketitle
\begin{abstract}
Cilia are hairlike structures protruding from nearly every cell in the body. Diseases known as \textit{ciliopathies}, where cilia function is disrupted, can result in a wide spectrum of disorders. However, most techniques for assessing ciliary motion rely on manual identification and tracking of cilia; this process is laborious and error-prone, and does not scale well. Even where automated ciliary motion analysis tools exist, their applicability is limited. Here, we propose an end-to-end computational machine learning pipeline that automatically identifies regions of cilia from videos, extracts patches of cilia, and classifies patients as exhibiting normal or abnormal ciliary motion. In particular, we demonstrate how convolutional LSTM are able to encode complex features while remaining sensitive enough to differentiate between a variety of motion patterns. Our framework achieves 90\% with only a few hundred training epochs. We find that the combination of segmentation and classification networks in a single pipeline yields performance comparable to existing computational pipelines, while providing the additional benefit of an end-to-end, fully-automated analysis toolbox for ciliary motion. 
\end{abstract}
\vspace{-1em}
\section{Introduction}
\vspace{-1em}
\label{sec:intro}

Cilia are microtubule based hair-like projections of the cell that can be motile or immotile, and in humans are found on nearly every cell of the body. Ciliopathies, or diseases with disruption of nonmotile or motile cilia function, can result in a wide spectrum of disorders, ranging from sinopulmonary disease such as in primary ciliary dyskinesia (PCD)~\cite{o2007diagnosing}, to mirror symmetric organ placement or randomized left-right organ placement as in heterotaxy~\cite{garrod2014airway}. Each of these conditions are associated with increased respiratory complications and poor postsurgical outcomes~\cite{nakhleh2012high, harden2013increased}. Diagnosing patients with ciliary motion (CM) abnormalities prior to surgery may provide the clinician with opportunities to institute prophylactic respiratory therapies to prevent these complications. Together, these findings suggest motile cilia dysfunction may have a very broad clinical impact.

Current methods for assessing CM rely on a combination of techniques often used in concert, including electron microscopy~\cite{CiliaEM}, ciliary beat frequency (CBF)~\cite{olm2011primary, mantovani2010automated},  and visual assessment of ciliary beat pattern by expert reviewers~\cite{o2012analysis,CiliaCM}. However, each of these methods has drawbacks and limitations~\cite{QuinnSTM}; in addition, none are amenable to cross-institutional comparisons and collaborations. Some semi-automated methods have been proposed~\cite{QuinnSTM,leigh2017assessment}, but these are all of limited utility to clinicians, requiring some form of manual annotation.

To overcome these deficiencies, we have developed an end-to-end computational pipeline using deep neural networks. Deep learning approaches have attained state of the art performance on many benchmark datasets in biomedical imaging, and are ideal for spatiotemporal analysis. Our pipeline automatically identifies regions of high-speed digital videos of ciliary biopsies which contain \textit{beating} or \textit{non-beating} cilia, and extracts them for downstream analysis. Once the regions of cilia are extracted, the temporal behavior of the cilia is parameterized and used to train a binary classifier. Finally, the classifier predicts the motion of the cilia for each patient. Once the stacked deep nets are trained, videos of ciliary biopsies can be fed directly into our pipeline and a CM prediction rendered, without requiring any manual intervention from clinicians.

\vspace{-1em}
\section{Data}
\vspace{-1em}
\label{sec:FCDN}

We used data from our previous work~\cite{QuinnSTM} that includes: nasal brush biopsies from 75 patients (35 healthy controls, 40 with a diagnosed ciliopathy), totaling 268 videos. Nasal epithelial tissue was collected by curettage of the inferior nasal turbinate under direct visualization with an appropriately sized nasal speculum using Rhino-probe (Arlington Scientific). Three passages were made, and the collected tissue was resuspended in L-15 medium (Invitrogen) for immediate videomicroscopy using a Leica inverted microscope with a 100$\times$ oil objective and differential interference contrast optics. Digital high-speed videos were recorded at a sampling frequency of 200 Hz using a Phantom v4.2 camera. To establish ground truth CM, these samples were analyzed by a panel of researchers blinded to the subject's clinical diagnosis and associated pathology reports. After reviewing all videos associated with a patient, a call of normal or abnormal CM was made by consensus. Where differences could not be resolved, the majority vote was accepted.

\vspace{-1em}
\section{Methods}
\vspace{-1em}
\label{sec:MTHD}

\subsection{Data Preprocessing and Augmentation}

Each video depicted cilia combined with varying levels of recording artifacts such as extraneous camera movement, uneven lighting, or poor focus. Additionally, cilia could be depicted at different angles relative to the camera's perspective, drastically altering appearance. To address this, we created four annotation ``classes'' in the first of our stacked deep networks: side-view (lateral) cilia, top-down cilia, cell body, and background. These four annotation classes were used for creating ground-truth segmentation masks using ITK-SNAP for a small number of videos. Random crops of 256 $\times$ 256 and horizontal vertical flips were used for data augmentation. Some other videos were discarded due to poor quality recording and absence of discernible cilia.

After identifying spatial regions containing cilia, we computed spatial and temporal derivatives of the optical flow~\cite{sun2010secrets} to derive differential invariants: instantaneous rotation, divergence, and deformation. In this study, we used only rotation; this quantity is a linear feature transformation that has demonstrated good empirical performance for differentiating CM patterns~\cite{QuinnSTM,quinn2011novel}. We extracted small patches from the regions predicted to contain cilia, and fed the corresponding rotation values at these patches coordinates into our second stacked deep network to build higher-order features for classification.

\vspace{-0.5em}
\subsection{DenseNets for Cilia Segmentation}
\vspace{-0.5em}
The first step in our end-to-end pipeline is the automated identification of regions in a video containing cilia. We motivate the use of densely-connected convolutional networks, or ``DenseNets,'' to automatically generate semantic segmentations of a given input video.

DenseNets entail the use of fully-connected neural networks where each layer has direct access to the gradients and loss from the original input. While densely-connected layers add more parameters per layer, the overall number of parameters is reduced, as fewer feature maps are needed in each layer. This allows DenseNets to be very deep while remaining parameter-efficient. This particular architecture is highly amenable to image-related tasks, such as semantic segmentation, where the fully-connected layers can propagate high-resolution information regarding where one region ``ends'' and another ``begins''; in our case, this is ideal for automatically differentiating regions containing cilia from those that do not, \textit{regardless of whether the cilia are moving.} We implement a version of fully convolutional dense networks for segmentation similar to the Tiramisu architecture~\cite{Tiramisu}.

\begin{figure}
\centering
\includegraphics[width=7cm, height=8.5cm]{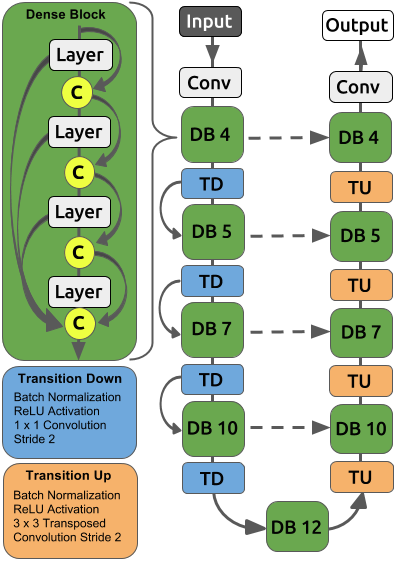}
\caption{Diagram of FC-DenseNet 74, so-named for having 74 total layers. Yellow circles represent concatenation operation and dashed lines represent skip connections.}
\label{fig:dn}
\end{figure}

The DenseNet architecture is composed of dense blocks (DB) in both the downsampling and upsampling paths (Fig.~\ref{fig:dn}), with multiple layers stacked in each DB. Ultimately a bottleneck is reached, after which the number of layers decrease in each subsequent DB~\cite{DenseNet}. Skip connections connect DBs in the downsampling and upsampling paths to facilitate information flow from shallow layers so high-level features can be reused in deep layers. A single convolutional layer is added before the first block and after the last block. Ultimately, the network builds a sophisticated set of feature maps that, given a ground-truth segmentation map, will predict masks for semantic regions of new, unobserved input image. 

\vspace{-0.5em}
\subsection{Convolutional LSTMs for CM Classification}
\vspace{-0.5em}

The second step in our end-to-end pipeline is modeling the CM as time series. While previous work has demonstrated some promise using Markov chains and autoregressive models~\cite{QuinnSTM}, and even using simple 3D convolutional networks to capture three time points simultaneously in a single deep network~\cite{lu2017classification}, we propose to use recurrent neural networks (RNNs) to leverage much longer-term temporal dependencies in the data, theoretically enabling much deeper and richer representations of CM.

We use a variant of RNNs called ``long short-term memory'' (LSTM) networks, which use a series of gates to intelligently determine what portions of input sequences should be ``remembered'' and which should be ``forgotten.'' These gates include the \textit{forget} (Eq.~\ref{eq:forget}), \textit{input} (Eq.~\ref{eq:input}), and \textit{output} (Eq.~\ref{eq:output}) gates, which determine whether a new input should be incorporated into the neuron's existing state (Eq.~\ref{eq:update}) or thrown away entirely; this enables the neuron to learn long-term dependencies. For sequences of images, such as videos of cilia biopsies, convolutional LSTM networks are ideal. Convolutional LSTMs~\cite{convLSTM} are similar to standard LSTM networks, except the inputs to each gate inside an LSTM neuron are convolved through kernel filters to extract spatial features, exactly like in a standard convolutional layer. These convolutional gates also preserve the temporal information inside the LSTM. See ~\cref{eq:forget,eq:input,eq:output,eq:cell,eq:update} for the convolutional LSTM transformations at each gate ($\circ$ denotes entry-wise product and  $*$ denotes convolution).

\vspace{-1.5em}
\begin{align}
f_t &= \sigma(W_{f} * x_t + U_{f} * h_{t-1} + V_{f} \circ c_{t-1} + b_f) \label{eq:forget} \\
i_t &= \sigma(W_{i} * x_t + U_{i} * h_{t-1} + V_{i} \circ c_{t-1} + b_i) \label{eq:input} \\
o_t &= \sigma(W_{o} * x_t + U_{o} * h_{t-1} + V_{o} \circ c_{t-1} + b_o) \label{eq:output} \\
c_t &= f_t \circ c_{t-1} + i_t \circ \sigma(W_{c} * x_t + U_{c} * h_{t-1} + b_c) \label{eq:cell} \\
h_t &= o_t \circ \sigma_h(c_t) \label{eq:update}
\end{align}
\vspace{-1.5em}

\begin{figure}
\begin{tabular}{ll}
\includegraphics[trim={3cm 0 3cm 0}, clip, width=4cm, height=4cm]{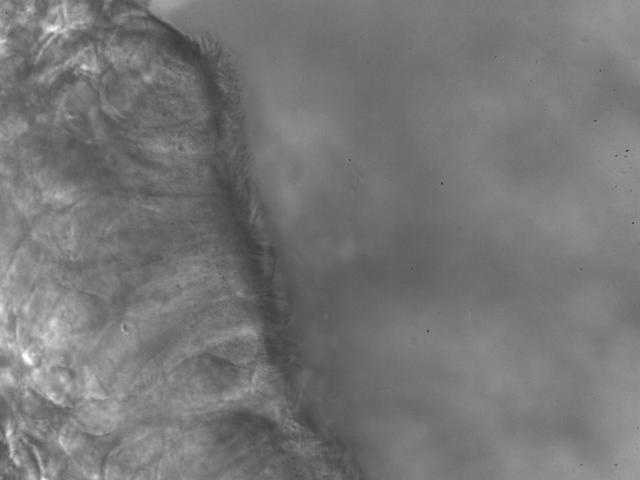}
\includegraphics[trim={3cm 0 3cm 0}, clip, width=4cm, height=4cm]{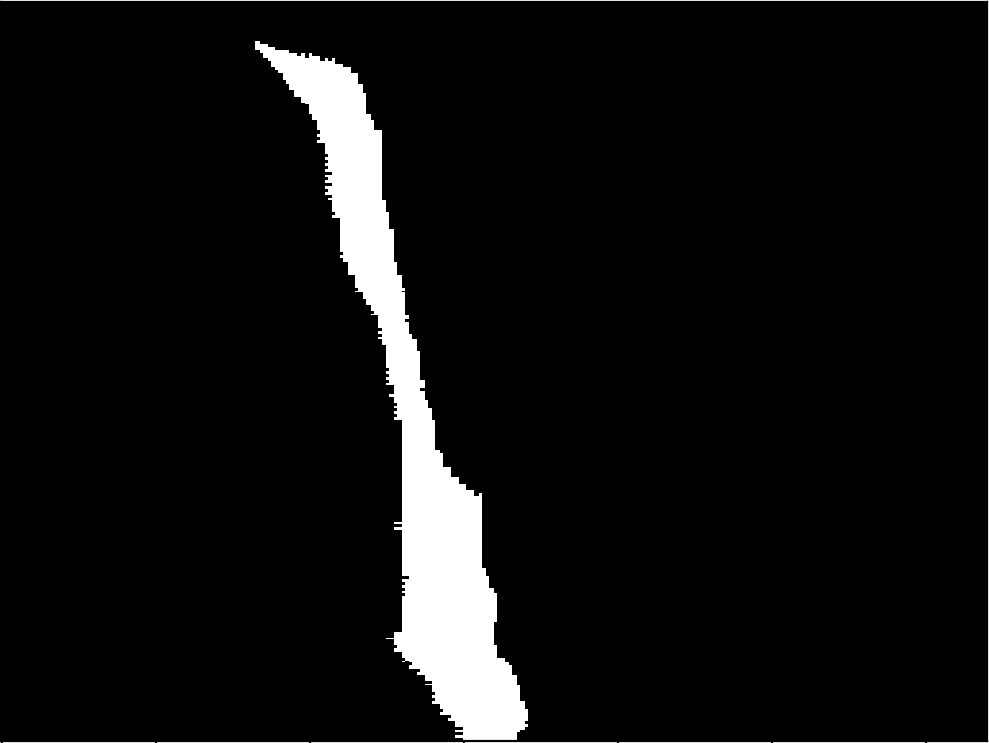}
\\
\includegraphics[trim={3cm 0 3cm 0}, clip, width=4cm, height=4cm]{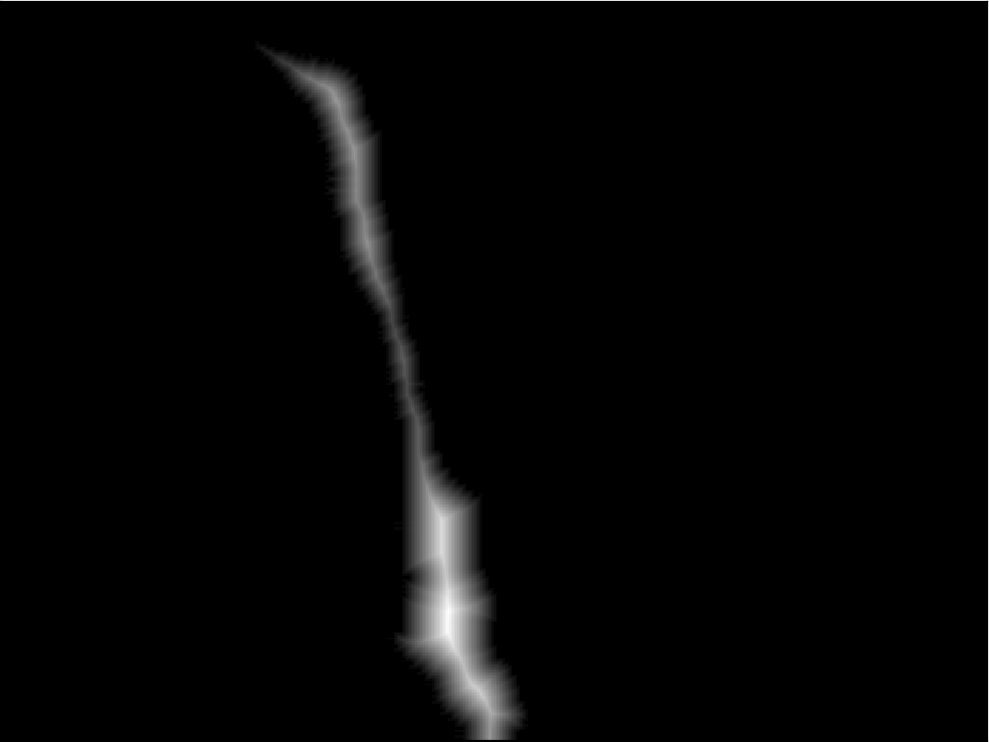}
\includegraphics[trim={3cm 0 3cm 0}, clip, width=4cm, height=4cm]{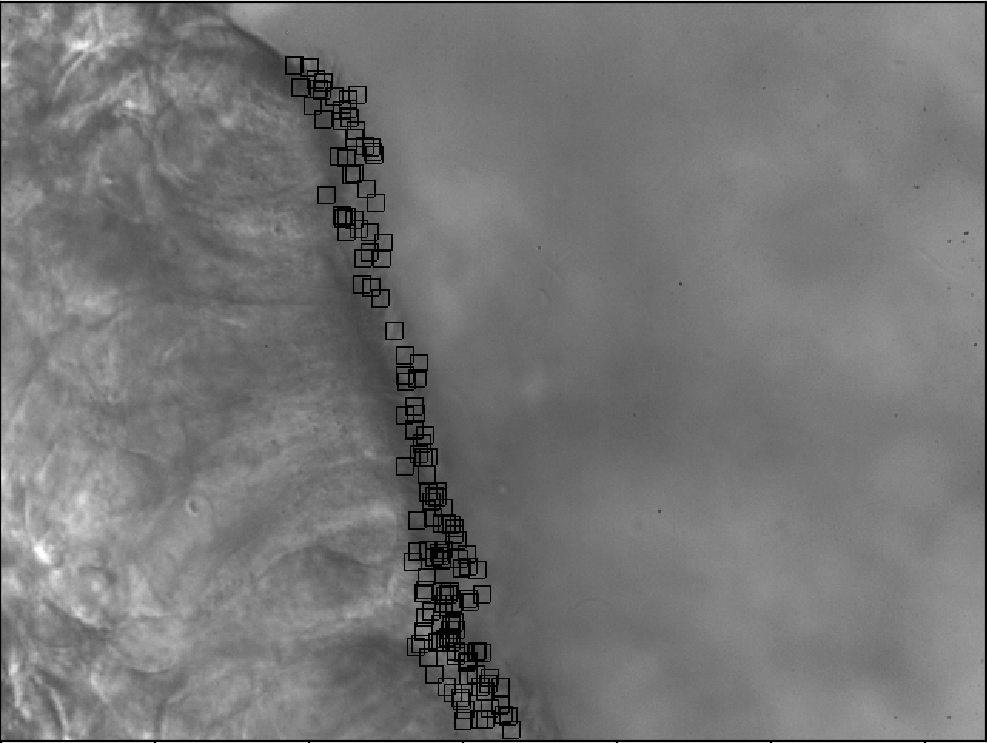}
\end{tabular}
\caption{Inputs and outputs of the pipeline. Input videos (upper left) are segmented, producing a segmentation mask (upper right). From this mask, a probability map is computed (bottom left) to sample patches (bottom right) that are then fed to the convolutional LSTM for analysis and classification.}
\label{fig:masks}
\end{figure}

\vspace{-1em}
\section{Results}
\vspace{-1em}

For cilia segmentation, we trained DenseNets with different numbers of layers to study the optimal tradeoff between speed and accuracy. We trained with the Adam optimizer to minimize categorical cross-entropy loss and vary regularization parameters tunings: dropout, $l^2$ weight decay, and learning rate annealing. Each model was trained for 100 epochs with a batch size of 4 on $2  \times$ Titan X GPU cards. All networks were implemented in Tensorflow with Keras. We evaluated resulting models on overall pixel classification accuracy, using the class with the highest probability for each pixel as the predicted pixel class. 
%

We found that DenseNets with 74 layers and 2.4 million parameters struck a good balance between accuracy and training time. The model produced quality segmentation maps of cilia, with a weighted Dice coefficient of 0.437 for cilia mask predictions (Table~\ref{table:seg}), beating out the seminal U-Net architecture~\cite{U-Net} in accuracy, and with an order of magnitude fewer parameters.

\setlength{\tabcolsep}{3pt}
\begin{table}
	\centering
	\begin{tabular}{ c | c | c | c | c}
		\hline
		Model & Parameters & Dropout & Decay & Accuracy \\ \hline
		\hline
		U-Net~\cite{U-Net} & 30 M & 0.3 & 0.001 & 76.9\% \\ 
		FC-DenseNet 55 & 2.4 M & 0.5 & 1$e^{-3}$ & 81.1\% \\ 
		FC-DenseNet 79 & 5 M & 0.3 & 1$e^{-4}$ & 84.7\% \\ 
		FC-DenseNet 109 & 9.4 M & 0.1 & None & 86.2\% \\
	\end{tabular}
	\caption{Segmentation performances of models using densely connected layers. Parameters are represented in millions.}
	\label{table:seg}
\vspace{-1.5em}
\end{table}

From the segmentation masks of the first stack of deep networks (Fig.~\ref{fig:masks}), we extract small patches. We observed from the segmentation results that the cilia were most likely to be found in the middle of the predicted masks. Therefore, we computed a distance map from the mask (Fig.~\ref{fig:masks}, bottom left) and used this as a sampling distribution. We sampled from the pixels within this mask without replacement until a saturation threshold was reached (proportional to the ratio of the area of the mask to the size of the patches). We then used the coordinates of the sampled pixels as the centers of $11 \times 11$ patches, and extracted 250 frames for each patch from the rotation data.

We collected a total of 24,577 patches from 75 patients. The label of each patch (normal or abnormal) was inherited from the patient. We performed cross-validation with three splits of 7,898, 8,519, and 8,160 patches (patches from the same patient were retained within the same fold to prevent testing contamination). Random horizontal and vertical flip augmentations of patches were performed during training. We used a convolutional LSTM with binary softmax classifier. We trained for 200 epochs with early stopping, using binary cross-entropy as the loss function.

For each patch in the validation set, the output probabilities from the final softmax layer was rounded based on a threshold of 0.5. All patches from the same video were averaged and rounded to 1 if above 0.5, and 0 if below 0.5. The resulting predicted classes for each video voted with a simple majority to determine the overall label of the patient. The classifier had an overall validation accuracy of 88\% and with an F1 score of 0.8965 (Table~\ref{table:acc}). It performed extremely well on cilia depicting abnormal CM (Table~\ref{table:conf}).

\begin{table}[h]
\centering
\begin{tabular}{ c | c | c |c | c }
	\hline
	Epochs & Accuracy & F1 & Recall & Precision \\ \hline
	\hline
	100 & 0.81 & 0.84 & 0.93 & 0.77 \\ \hline
	200 & 0.88 & 0.90 & 0.98 & 0.83 \\ 
\end{tabular}
\caption{Convolutional LSTM classification performance as a function of training epochs.}
\label{table:acc}
\end{table}

\begin{table}[h]
\centering
\begin{tabular}{ c | c | c}
	& Normal (predicted) & Abnormal (predicted) \\ \hline
	Normal (actual) & 27 & 8 \\ \hline
	Abnormal (actual) & 1 & 39 \\ 
\end{tabular}
\caption{Patient classification confusion matrix.}
\label{table:conf}
\end{table}


\vspace{-1em}
\section{Discussion}
\vspace{-1em}

The convolutional LSTM classifies abnormal patients nearly perfectly; it struggles somewhat more on patients with normal CM. We found a similar classifier pathology in~\cite{QuinnSTM}, despite using a distinct classification pipeline, suggesting the ``distributions'' of normal and abnormal CM overlap in that abnormal is a highly restricted subset of normal. Also, we observed mask predictions with in videos with unambiguous cell bodies had much better classification results. Upon inspection, we found the masks in these cases were of much higher quality and more likely to contain cilia, and therefore the extracted patches as well, directly impacting classification.

\vspace{-1em}
\section{Conclusions and Future Work}
\vspace{-1em}
In this paper, we demonstrate the efficacy of a pipeline of stacked deep nets for fully-automated, end-to-end analysis of CM. While achieving a high level of accuracy, future work will entail deeper training of the segmentation model to incorporate temporal information when determining the masks. Additionally, given the high variability in normal CM, we eventually aim to conduct fully unsupervised CM analysis using unmixing to determine motion subtypes and their pathological implications.

\vspace{-1em}
\section{Acknowledgments}
\vspace{-1em}
The authors gratefully acknowledge the support of NVIDIA Corporation with the donation of the Titan X Pascal GPU used for this research.

\vspace{-1em}
\small{
\bibliographystyle{IEEEbib}
\bibliography{refs}
}

\end{document}